\documentclass[letterpaper, 10 pt, conference]{ieeeconf}  

\IEEEoverridecommandlockouts                              

\usepackage[T1]{fontenc}

\usepackage{subfigure}
\usepackage{float}

\usepackage{graphicx}
\graphicspath{{./images/}}

\usepackage{amsmath}
\usepackage[table, xcdraw]{xcolor}

\usepackage{cite}

\usepackage{hyperref}

\title{\LARGE \bf
Monocular Depth Estimation and Segmentation for Transparent Object with Iterative Semantic and Geometric Fusion
}

\author{Jiangyuan Liu$^{1,2,4}$, Hongxuan Ma$^{1,2}$, Yuxin Guo$^{1,2}$, Yuhao Zhao$^{1,2}$, Chi Zhang$^{3}$, Wei Sui$^{4}$, Wei Zou$^{1,2\dagger}$
\thanks{$^{1}$School of Artificial Intelligence, University of Chinese Academy of Sciences}
\thanks{$^{2}$State Key Laboratory of Multimodal Artificial Intelligence Systems (MAIS), Institute of Automation of Chinese Academy of Sciences}%
\thanks{$^{3}$School of Information Science and Technology, Shijiazhuang Tiedao University}%
\thanks{$^{4}$D-Robotics}%
\thanks{$^{\dagger}$Corresponding to {\tt\small wei.zou@ia.ac.cn}}%
}

\begin{document}

\maketitle
\thispagestyle{empty}
\pagestyle{empty}

\begin{abstract}


Transparent object perception is indispensable for numerous robotic tasks. However, accurately segmenting and estimating the depth of transparent objects remain challenging due to complex optical properties. Existing methods primarily delve into only one task using extra inputs or specialized sensors, neglecting the valuable interactions among tasks and the subsequent refinement process, leading to suboptimal and blurry predictions. To address these issues, we propose a monocular framework, which is the first to excel in both segmentation and depth estimation of transparent objects, with only a single-image input. Specifically, we devise a novel semantic and geometric fusion module, effectively integrating the multi-scale information between tasks. In addition, drawing inspiration from human perception of objects, we further incorporate an iterative strategy, which progressively refines initial features for clearer results. Experiments on two challenging synthetic and real-world datasets demonstrate that our model surpasses state-of-the-art monocular, stereo, and multi-view methods by a large margin of about 38.8\%-46.2\% with only a single RGB input. Codes and models are publicly available at \href{https://github.com/L-J-Yuan/MODEST}{https://github.com/L-J-Yuan/MODEST}.

\end{abstract}

\section{INTRODUCTION}

Transparent objects such as bottles, flasks, and windows are ubiquitous in various domains, like laboratories, industries, or daily life. For robots in these scenarios, accurately detecting and estimating the depth of transparent objects are usually prerequisites for subsequent manipulation and navigation tasks \cite{c1}. However, transparent objects often lack clear texture and blend with the background in most RGB images, due to their complex refraction and reflection characteristics \cite{c2}. Additionally, commercial depth cameras also struggle to perceive such objects \cite{c7}, thus producing incomplete and noisy depth maps. These failures of conventional sensors hinder the development of downstream tasks like grasping.

\begin{figure}[!tb]
    \centering
	\includegraphics[width=3.5in]{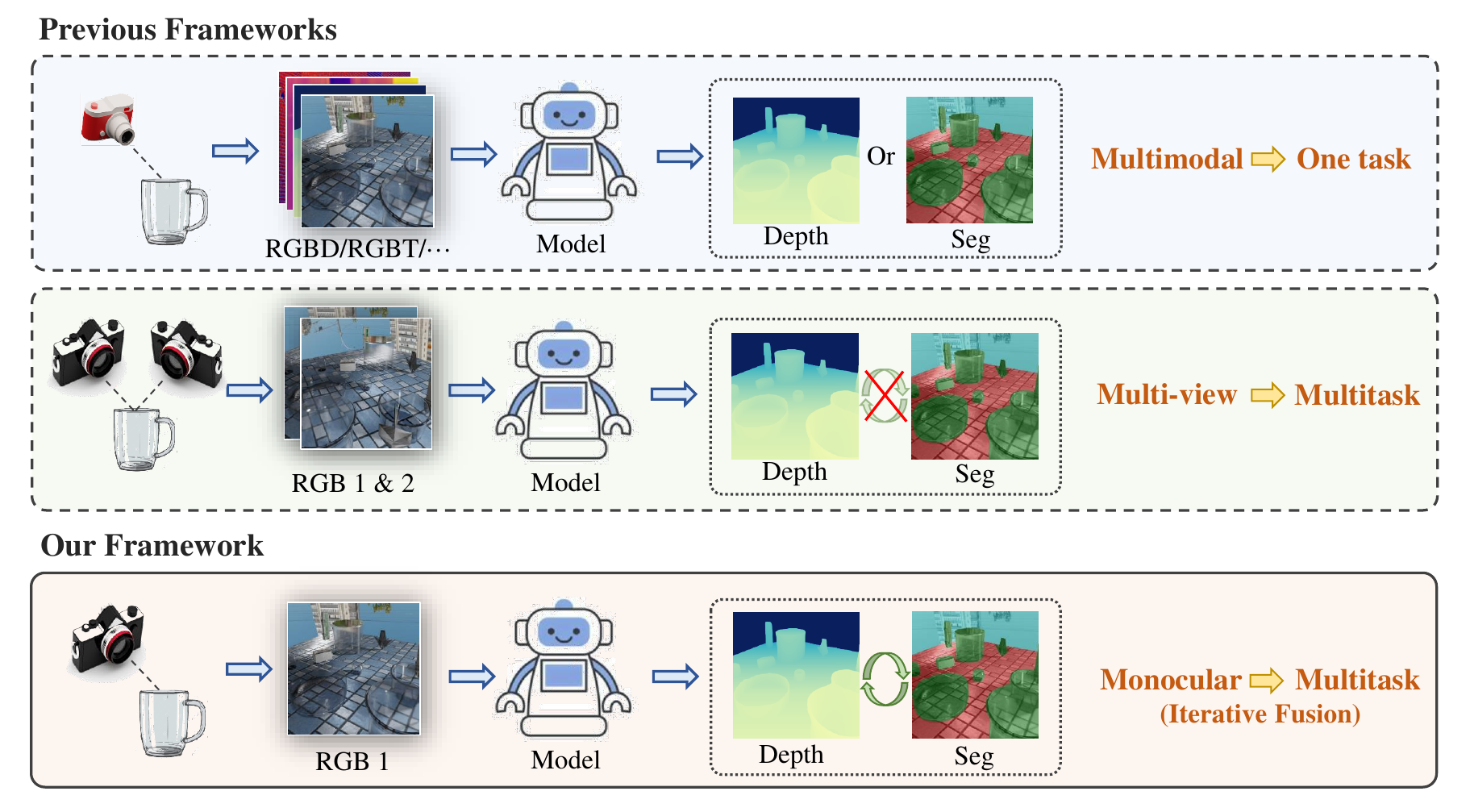}
    \caption{Previous frameworks rely either on multi-view inputs or additional modalities (e.g., depth maps, thermal images) to make predictions. Differently, we propose the first monocular framework that utilizes iterative cross-task fusion to improve both depth and segmentation performance. }
    \label{framework}
    \vspace{-8pt}
\end{figure}

Confronting these issues, previous researches independently focus on either segmentation or depth estimation of transparent objects, using supplementary modalities, as shown in Fig. \ref{framework}. For example, some works resort to specialized sensors, such as polarized camera \cite{c5}, RGB-Thermal camera \cite{c6}, etc., which are typically expensive and difficult to obtain. Other methods utilize multi-view RGB images \cite{c11,c12,c13} or additional depth maps \cite{c8,c9,c10} as inputs, leading to substantial time overhead and suboptimal performance due to incomplete and noisy depth maps of transparent objects. More recently, SimNet \cite{c14} and MVTrans \cite{c15} adopt multi-task frameworks for transparent object perception in stereo and multi-view settings, respectively. 
However, they overlook the beneficial information and interactions across multiple tasks, resulting in notably inferior and unbalanced performance of both segmentation and depth estimation.

To address these issues, we analyze and identify two key breakthroughs. \textbf{(a) Integrating semantic and geometric interactions into complementary tasks can fully exploit the useful mutual information.} As an ill-posed problem, monocular depth estimation is particularly challenging for transparent objects. Fortunately, semantic segmentation is informative for depth estimation by offering semantic and contextual clues \cite{c17}. Similarly, depth estimation could also provide valuable multi-scale geometric information for segmentation, such as boundaries, surfaces, and shapes to assist in determining semantic categories \cite{c18}. \textbf{(b) Iterating multi-scale fusion continuously can refine the initial fusion results.} When humans observe inconspicuous objects, we tend to notice the overall outline of the object first, then the local details \cite{c43}. Inspired by this, we believe that updating features in a coarse-to-fine fashion facilitates transparent object perception.
 
Based on our analysis, we for the first time propose a monocular framework to concurrently predict precise segmentation and depth for transparent objects. Different from previous works, we take the simplest and most efficient form using only a single RGB input, as shown in Fig. \ref{framework}. Specifically, to fully exploit the complementary information across tasks, we design a novel semantic and geometric fusion module that adaptively interacts with the features of both tasks, allowing the model to effectively enhance the predictions, especially for depth estimation. Moreover, to obtain more fine-grained and accurate predictions, we propose an iterative strategy to repeatedly update the initial features through a shared decoder, thereby further improving the performance of both tasks. Extensive experiments on both synthetic and real-world datasets show that, our model is superior to general multi-task methods, and outperforms state-of-the-art stereo and multi-view methods significantly in both depth and segmentation for transparent objects.

In summary, our main contributions are as follows:

\begin{itemize}

\item To the best of our knowledge, we propose the first end-to-end monocular framework excelling in predicting both depth and segmentation for transparent objects.
\item The key advantages of our approach lie in the semantic and geometric fusion module and an innovative iterative strategy, which better leverage the complementary information between the two tasks, significantly improving transparent object perception.
\item Experimental results demonstrate that our model outperforms state-of-the-art monocular and even multi-view methods by a large margin quantitatively and qualitatively, on both synthetic and real datasets.

\end{itemize}

\section{RELATED WORK}

\subsection{Transparent Object Segmentation}

Accurate detection or segmentation is usually the first step in perceiving and manipulating untextured transparent objects. On the one hand, many existing works utilize specific visual cues to segment transparent objects. For instance, TransLab \cite{c3} and EBLNet \cite{c4} demonstrated the effectiveness of boundaries for locating transparent objects. GDNet \cite{c21} and RFENet \cite{c22} proposed novel feature fusion modules to enhance performance by better utilizing contextual and reciprocal features, respectively. On the other hand, some methods obtain additional information gains by means of different input modalities. PGSNet \cite{c5} employed a polarized camera to extract optical cues beneficial for segmentation. In \cite{c6}, thermal images were combined by a multi-modal fusion module to assist in detecting glass surfaces. Differently, our method only takes a single RGB image as input, without relying on additional modalities.

\subsection{Transparent Object Depth Estimation}

The techniques for depth estimation of transparent objects can be roughly classified into depth completion and NeRF-based methods. ClearGrasp \cite{c7} pioneered the use of RGB-D input for transparent object depth completion. Successive improvements have come from LIDF-Refine \cite{c23} and TransparentNet \cite{c24} by lifting depth maps to point clouds and performing completion. A more recent work TODE \cite{c10} leveraged swin transformer \cite{c25} to better capture the global information. Following recent advancements in NeRF \cite{c16}, DexNeRF \cite{c11} and EvoNeRF \cite{c12} employed implicit functions to represent transparent objects, though the optimization processes were time-consuming. GraspNeRF \cite{c13} and ResidualNeRF \cite{c44} later sped up inference by utilizing the generalizable NeRF and decoupling the background, respectively. Most methods predict depth only once, while we take an iterative way for further refinement.

\subsection{Multi-task Predictions for Transparent Objects}

Multi-task dense predictions aim to learn multiple tasks jointly in a unified framework \cite{c42,c28,c33}. ClearGrasp \cite{c7} adopted edges, masks and surface normals as intermediate representations for optimizing depth. SimNet explored a multi-task framework based on stereo input to support transparent object manipulation, while recently MVTrans \cite{c15} extended it by introducing multiple views. However, none of the above methods for transparent objects leveraged inter-task interactions. In contrast, we propose a fusion module to fully exploit the complementary information between different tasks.





\begin{figure*}[!htb]
    \centering
    \includegraphics[width=7in]{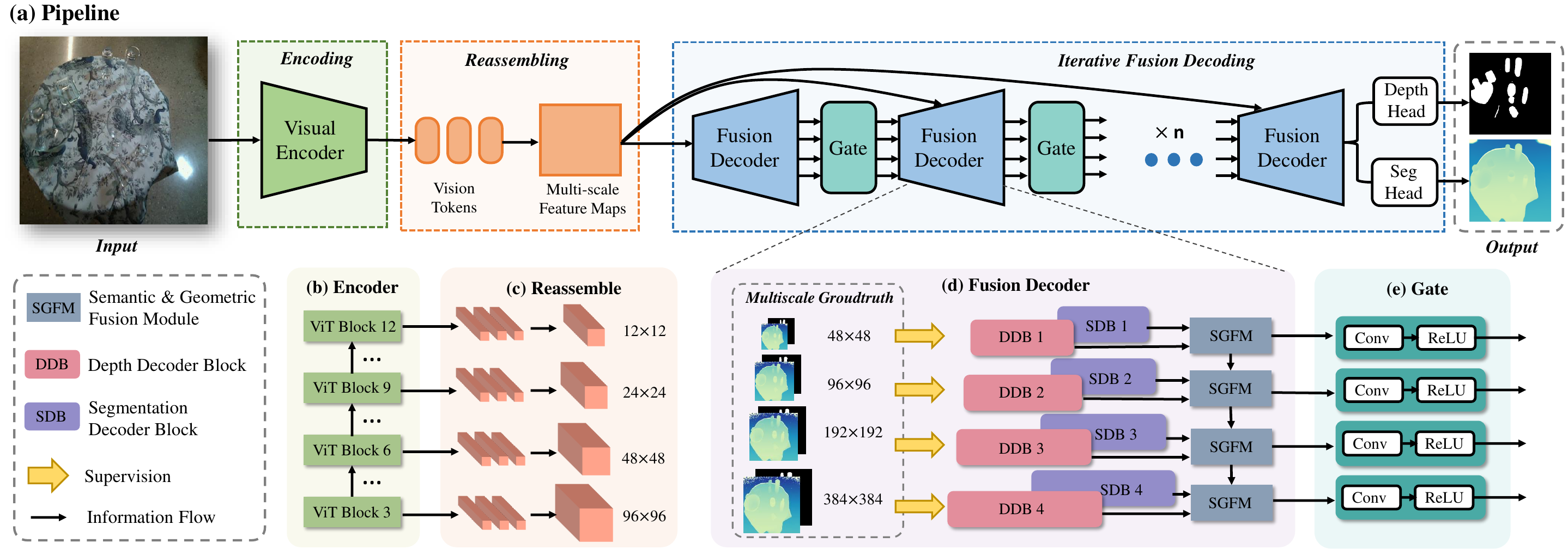}
    \caption{\textbf{ Overview of our proposed end-to-end framework.} (a) Given an RGB input, our model jointly predicts depth and segmentation mask through encoding, reassembling, and iterative fusion decoding. (b) The encoder uses ViT \cite{c38} to extract vision tokens of four layers. (c) Then in the reassemble module, the tokens are transformed into multi-scale feature maps, forming two pyramids for depth and segmentation, respectively. (d) A novel semantic and geometric fusion module is designed in the decoder for better leveraging the complementary information of both tasks. (e) The shared-weight decoder is updated iteratively by lightweight gates to gradually refine the initial results. Final predictions are obtained by two heads after the last iteration.}
    \label{model}
    \vspace{-8pt}
\end{figure*}

\section{METHOD}

\subsection{Problem Statement and Method Overview} 

Given a single RGB image $ I \in R^{3 \times H \times W} $, where $H$ is the height and $W$ is the width of the image, the objective is to obtain an accurate segmentation mask $ S \in R^{N \times H \times W} $ and a depth map $ D \in R^{H \times W} $ for transparent objects, where $N$ is the number of semantic categories. Our model learns a function $ f $ that maps the input to two outputs, defined as $ (S, D) = f(I) $.

As depicted in Fig. \ref{model}, the overall architecture is composed of a transformer-based encoder, a reassemble module and an iterative fusion decoder. In the encoder, the input RGB image is first processed and passed through multiple transformer blocks to extract features as vision tokens. Then we assemble tokens from different layers into multi-scale feature maps, which form two feature pyramids for depth and segmentation, respectively. In the decoder, the two branches are merged together through our semantic and geometric fusion module. This multi-scale fusion and decoding process is iteratively refined through gated units several times to obtain the final depth and segmentation predictions.

\subsection{Transformer Encoder}

Existing methods dealing with transparent objects mostly utilize CNN as feature extractors \cite{c7,c15}. However, we argue that compared with traditional convolution operators, attention mechanisms provide global contextual representation, which has proven to be especially effective for transparent objects \cite{c10}. Thus we employ the vision transformer (ViT) \cite{c38} as our backbone to extract multi-layer features. We first crop the input RGB image into non-overlapping patches, followed by a linear projection to embed the patches into tokens. Then the tokens are added by position embeddings and processed by multiple transformer blocks with multiheaded self-attention. The encoder consists of 12 transformer blocks, from which we select 4 layers of tokens, evenly distributed from shallow to deep, for the following module.

\subsection{Reassemble Module}

Since ViT encodes image features as tokens with the same spatial resolution, we need to convert them back to feature maps for subsequent fusion and prediction. Following DPT \cite{c39}, the vision tokens are reshaped into corresponding feature maps by concatenation and projection. To fully exploit features of different levels, we represent them in a multi-scale fashion, where deeper features correspond to smaller resolutions. The results of the reassemble module are two four-layer pyramids for depth and segmentation, respectively.

\subsection{Iterative Fusion Decoder}

In the decoder, the geometric features and semantic features from the two pyramids are integrated together with our proposed fusion module. Then we iteratively refine the features from the same shared-weight decoder through gated units to obtain more fine-grained predictions.

\begin{figure}[!tb]
    \centering
    \includegraphics[width=3in]{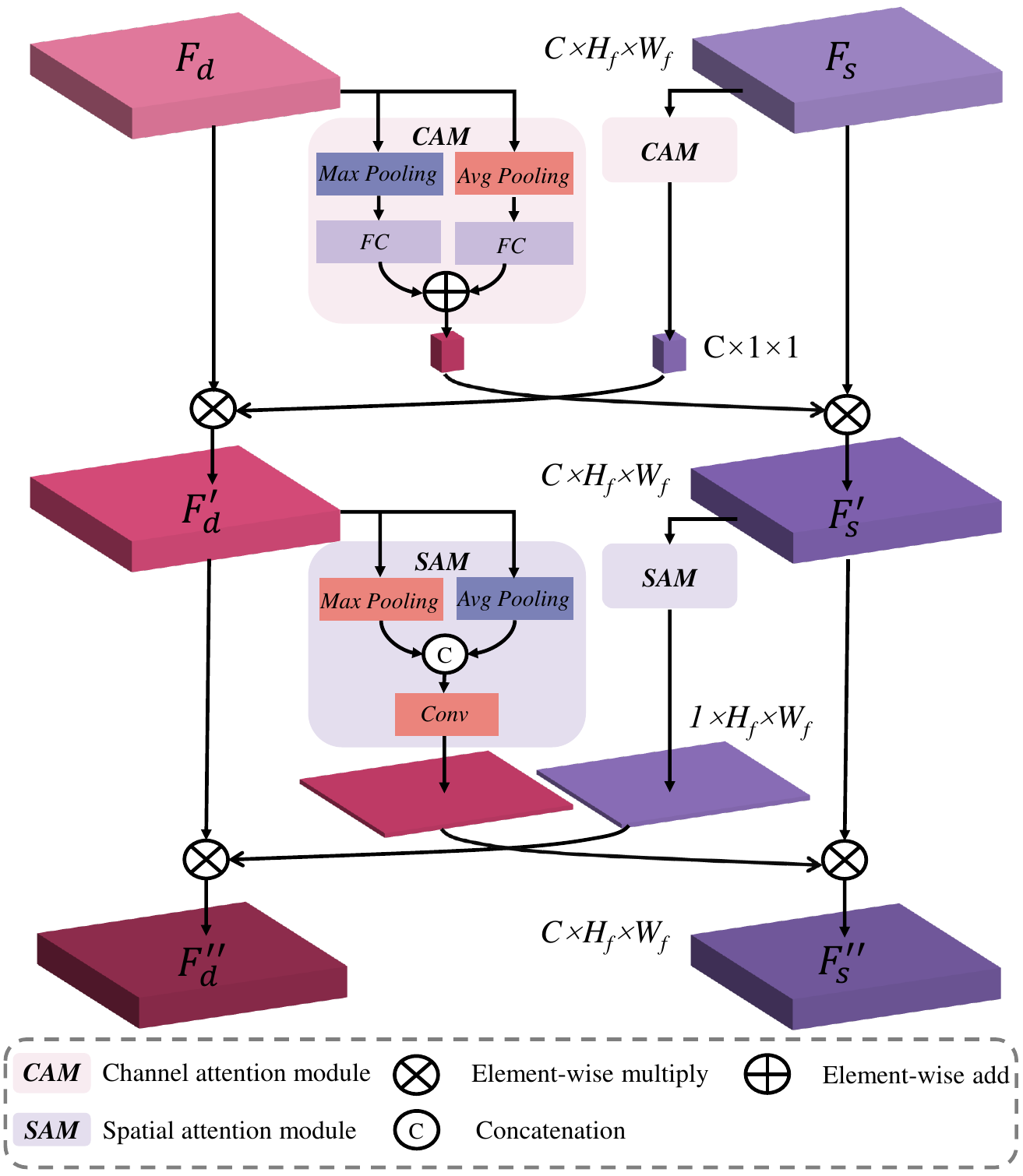}
    \caption{\textbf{Illustration of the semantic and geometric fusion module (SGFM).} $F_d$ and $F_s$ represent features of a certain layer of the depth and segmentation pyramid, respectively. The two feature maps are processed along both channel and spatial dimensions to adaptively emphasize semantic and geometric information. They are then cross-multiplied to achieve the fusion.}
    \label{module}
    \vspace{-8pt}
\end{figure}

{\bf Fusion Decoder.} Due to the optical properties of transparent objects, it is particularly difficult to predict depth and segmentation independently with a single RGB image \cite{c7}. To improve the performance of both tasks, inspired by \cite{c17}, we design a novel attention-based fusion module to fully exploit the complementary information of the two branches. With the two feature pyramids of depth and segmentation from the previous module, we apply semantic and geometric fusion at every layer to integrate multi-scale features. Without loss of generality, in Fig. \ref{module} we take one layer of the features as an example. Given a depth feature $F_{d} \in R^{C \times H_f \times W_F}$ and a segmentation feature $F_{s} \in R^{C \times H_f \times W_F}$ of a certain scale, we first apply a channel attention module and a spatial attention module to successively extract meaningful cues. Attention along channel and spatial axes has been proven to be effective in learning what and where to focus \cite{c40}. Leveraging this powerful representation ability, the module can automatically learn significant semantic and geometric information implied in depth and segmentation features, respectively. The informative extractions then interact with each other through symmetric multiplication. Concretely, the channel attention and spatial attention are computed as:
\setlength{\arraycolsep}{0.0em}
\begin{eqnarray}
CAM(F) &{}={}& \sigma(FC(AP(F)) + FC(MP(F))) \\
SAM(F) &{}={}& \sigma(c^{7 \times 7}[AP(F),MP(F)])
\end{eqnarray}
\setlength{\arraycolsep}{5pt}where $F$ represents either $F_{d}$ or $F_{s}$. $AP$ and $MP$ denote average pooling and max pooling, respectively. $FC$ denotes a fully connected layer. $c^{7 \times 7}$ represents a $7 \times 7$ convolution operation. $[\cdot , \cdot]$ denotes the concatenation operation and $\sigma$ is the sigmoid function.

Taking the depth feature as an example, the overall semantic and geometric fusion module is defined as:
\setlength{\arraycolsep}{0.0em}
\begin{eqnarray}
F_{d}^{'} &{}={}& F_{d} \otimes CAM(F_{s}) \\
F_{d}^{''} &{}={}& F_{d}^{'} \otimes SAM(F_{s}^{'})
\end{eqnarray}
\setlength{\arraycolsep}{5pt}where $\otimes$ denotes the element-wise multiplication. $F_{d}^{'}$ and $F_{s}^{'}$ are the intermediate representation of depth and segmentation features and $F_{d}^{''}$ represents the depth feature after fusion. Segmentation features are processed in the same way.


{\bf Iterative Refinement.} When faced with transparent objects, previous works with only one iteration of prediction tend to produce unclear results \cite{c14,c15}. To solve the problem, we instead propose an iterative refinement strategy to optimize depth and segmentation features in a coarse-to-fine manner. Taking the first multi-scale fusion results as the initial features, we update them repeatedly with a shared-weight decoder. The results from the previous iteration are passed to the next via lightweight gated units, which contain convolution operations and ReLu functions. The overall iterative process can be expressed as:
\begin{equation}
    F_n = f_d(F_e, Gate(F_{n-1}))
\end{equation}
where $F_{n-1}$ and $F_n$ are the set of all the multi-scale depth and segmentation features of iteration $n-1$ and $n$. $F_e$ denotes the features from the reassemble module and $f_d$ is the function represented by the shared decoder. Based on the features from the last layer and after the last iteration, two prediction heads consisting of convolutions and interpolations are adopted to obtain the final depth map and segmentation mask. To enforce the model to learn more details about transparent objects gradually, we apply multi-scale supervision from weak to strong to each iteration. The strength of each supervision is controlled by $n/N$, where $N$ is the total number of iterations and is set as 3 according to the ablation experiment.

\begin{figure*}[!tb]
    \centering
    \includegraphics[width=7in]{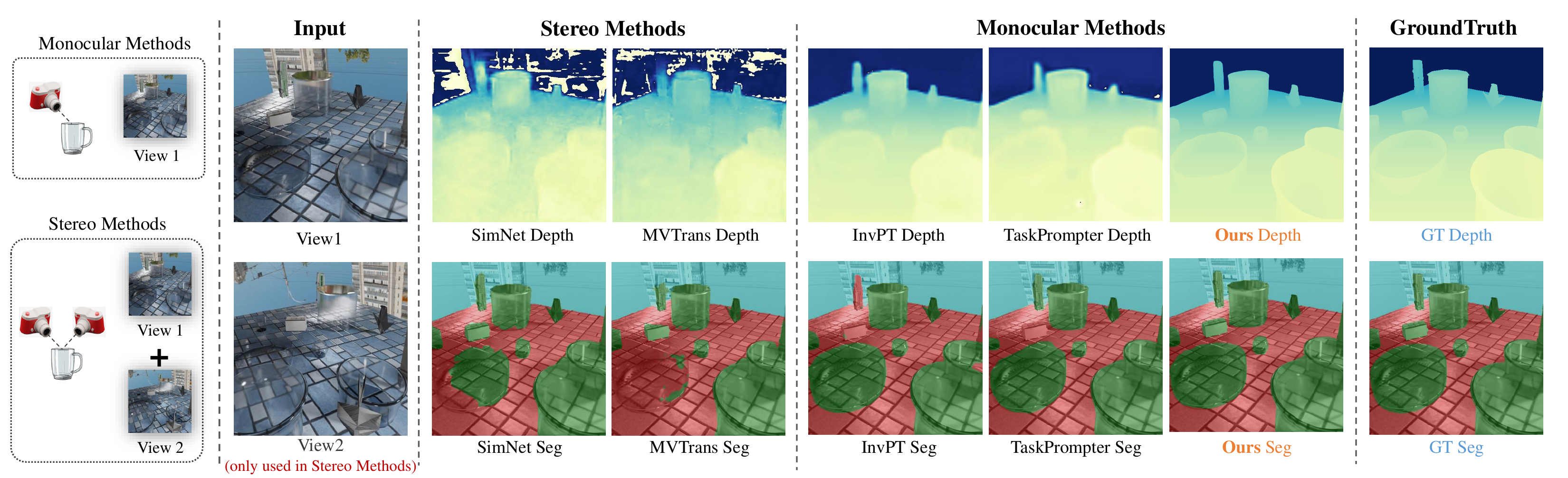}
    \caption{\textbf{Qualitative comparison on Syn-TODD dataset} of depth and segmentation, where Seg and GT stand for segmentation and ground truth, respectively. SimNet and MVTrans take both RGB images as input, while the other methods only take the first one as input. Obviously, our predictions are far better than all other methods with only single RGB as input.}
    \label{syntodd_fig}
    \vspace{2pt}
\end{figure*}

\begin{table*}[!htb]
    \renewcommand{\arraystretch}{1.1}
    \caption{Quantitative comparison of SOTA monocular, stereo, and multi-view methods on Syn-TODD dataset, where $\uparrow$ indicates that higher values are better and $\downarrow$ means that lower values are better. The last line shows the percentages by which our method exceeds the second-best results.}
    \label{table_syntodd}
    \centering
    \begin{tabular}{lccccccc}
    \hline
    \multicolumn{1}{l|}{}                                      & \multicolumn{1}{c|}{}                                             & \multicolumn{1}{c|}{}                                               & \multicolumn{3}{c|}{\textbf{Depth}}                                                                                                                                                     & \multicolumn{2}{c}{\textbf{Segmentation}}                                                   \\
    \multicolumn{1}{l|}{}                                      & \multicolumn{1}{c|}{Task}                                         & \multicolumn{1}{c|}{Modality}                                       & \multicolumn{1}{c|}{RMSE ($\downarrow$)}                                & \multicolumn{1}{c|}{MAE ($\downarrow$)}                                 & \multicolumn{1}{c|}{REL ($\downarrow$)}                                 & \multicolumn{1}{c|}{mAP ($\uparrow$)}                                 & IoU ($\uparrow$)                        \\ \hline
    \multicolumn{1}{l|}{InvPT \cite{c42}}                                 & \multicolumn{1}{c|}{general}                                      & \multicolumn{1}{c|}{monocular RGB}                                  & \multicolumn{1}{c|}{0.166}                                  & \multicolumn{1}{c|}{0.145}                                  & \multicolumn{1}{c|}{0.159}                                  & \multicolumn{1}{c|}{95.62}                                  & 89.74                         \\
    \multicolumn{1}{l|}{TaskPrompter \cite{c33}}                          & \multicolumn{1}{c|}{general}                                      & \multicolumn{1}{c|}{monocular RGB}                                  & \multicolumn{1}{c|}{0.247}                                  & \multicolumn{1}{c|}{0.233}                                  & \multicolumn{1}{c|}{0.247}                                  & \multicolumn{1}{c|}{{\underline{96.90}}}                            & {\underline{90.50}}                   \\ \hline
    \multicolumn{1}{l|}{SimNet \cite{c14}}                                & \multicolumn{1}{c|}{transparent}                                  & \multicolumn{1}{c|}{stereo RGB}                                     & \multicolumn{1}{c|}{1.229}                                  & \multicolumn{1}{c|}{1.020}                                  & \multicolumn{1}{c|}{0.975}                                  & \multicolumn{1}{c|}{48.21}                                  & 50.52                         \\
    \multicolumn{1}{l|}{MVTrans \cite{c15}}                               & \multicolumn{1}{c|}{transparent}                                  & \multicolumn{1}{c|}{2-view RGB}                                     & \multicolumn{1}{c|}{0.134}                                  & \multicolumn{1}{c|}{0.089}                                  & \multicolumn{1}{c|}{0.135}                                  & \multicolumn{1}{c|}{84.94}                                  & 79.52                         \\
    \multicolumn{1}{l|}{MVTrans}                               & \multicolumn{1}{c|}{transparent}                                  & \multicolumn{1}{c|}{3-view RGB}                                     & \multicolumn{1}{c|}{0.125}                                  & \multicolumn{1}{c|}{0.083}                                  & \multicolumn{1}{c|}{0.125}                                  & \multicolumn{1}{c|}{87.75}                                  & 81.89                         \\
    \multicolumn{1}{l|}{MVTrans}                               & \multicolumn{1}{c|}{transparent}                                  & \multicolumn{1}{c|}{5-view RGB}                                     & \multicolumn{1}{c|}{{\underline{0.124}}}                            & \multicolumn{1}{c|}{{\underline{0.080}}}                            & \multicolumn{1}{c|}{{\underline{0.117}}}                            & \multicolumn{1}{c|}{87.24}                                  & 81.30                         \\
    \rowcolor[HTML]{ECF4FF} 
    \multicolumn{1}{l|}{\cellcolor[HTML]{ECF4FF}\textbf{Ours}} & \multicolumn{1}{c|}{\cellcolor[HTML]{ECF4FF}\textbf{transparent}} & \multicolumn{1}{c|}{\cellcolor[HTML]{ECF4FF}\textbf{monocular RGB}} & \multicolumn{1}{c|}{\cellcolor[HTML]{ECF4FF}\textbf{0.070}} & \multicolumn{1}{c|}{\cellcolor[HTML]{ECF4FF}\textbf{0.052}} & \multicolumn{1}{c|}{\cellcolor[HTML]{ECF4FF}\textbf{0.068}} & \multicolumn{1}{c|}{\cellcolor[HTML]{ECF4FF}\textbf{97.83}} & \textbf{92.84}                \\ \hline
                                                               &                                                                   &                                                                     & {\color[HTML]{FE0000} +45.2\%}                              & {\color[HTML]{FE0000} +38.8\%}                              & {\color[HTML]{FE0000} +46.2\%}                              & {\color[HTML]{FE0000} +0.9\%}                               & {\color[HTML]{FE0000} +2.1\%} \\ \hline
    \end{tabular}%
\end{table*}

\begin{table}[!htb]
    \renewcommand{\arraystretch}{1.1}
    \caption{Comparison of multi-task methods on the ClearPose dataset.}
    \label{table_clearpose}
    \centering
    \begin{tabular}{l|ccc|cc}
    \hline
                           & \multicolumn{3}{c|}{\textbf{Depth}}                                            & \multicolumn{2}{c}{\textbf{Segmentation}}    \\
    \multicolumn{1}{c|}{}  & \multicolumn{1}{c|}{RMSE($\downarrow$)}     & \multicolumn{1}{c|}{MAE($\downarrow$)}      & REL($\downarrow$)      & \multicolumn{1}{c|}{mAP($\uparrow$)}      & IoU($\uparrow$)      \\ \hline
    InvPT                  & \multicolumn{1}{c|}{0.185}          & \multicolumn{1}{c|}{0.163}          & 0.212          & \multicolumn{1}{c|}{98.09}          & 85.91          \\
    TaskPrompter           & \multicolumn{1}{c|}{0.172}          & \multicolumn{1}{c|}{0.146}          & 0.190          & \multicolumn{1}{c|}{97.78}          & 85.00          \\ \rowcolor[HTML]{ECF4FF}
    \textbf{Ours} & \multicolumn{1}{c|}{\textbf{0.123}} & \multicolumn{1}{c|}{\textbf{0.081}} & \textbf{0.087} & \multicolumn{1}{c|}{\textbf{98.21}} & \textbf{86.27} \\ \hline
    \end{tabular}
\end{table}

\begin{table}[!htb]
    \renewcommand{\arraystretch}{1.1}
    \caption{Ablation study on the semantic and geometric fusion module. Depth only and Seg only indicate single-task predictions and w/o means without.}
    \label{table_ablation}
    \centering
    \resizebox{\columnwidth}{!}{%
    \begin{tabular}{l|ccc|cc}
    \hline
                   & \multicolumn{3}{c|}{\textbf{Depth}}                                                                                                        & \multicolumn{2}{c}{\textbf{Segmentation}}                                    \\
                   & \multicolumn{1}{c|}{RMSE($\downarrow$)}                                   & \multicolumn{1}{c|}{MAE($\downarrow$)}                                    & REL($\downarrow$)            & \multicolumn{1}{c|}{mAP($\uparrow$)}                                    & IoU($\uparrow$)            \\ \hline
Ours Depth Only    & \multicolumn{1}{c|}{0.080}                                  & \multicolumn{1}{c|}{0.061}                                  & 0.073          & \multicolumn{1}{c|}{-}                                      & -              \\
Ours Seg Only      & \multicolumn{1}{c|}{-}                                      & \multicolumn{1}{c|}{-}                                      & -              & \multicolumn{1}{c|}{97.12}                                  & 91.57          \\
Ours w/o SGFM      & \multicolumn{1}{c|}{0.087}                                  & \multicolumn{1}{c|}{0.064}                                  & 0.076          & \multicolumn{1}{c|}{96.17}                                  & 90.28          \\
\rowcolor[HTML]{ECF4FF} 
\textbf{Ours Full} & \multicolumn{1}{c|}{\cellcolor[HTML]{ECF4FF}\textbf{0.070}} & \multicolumn{1}{c|}{\cellcolor[HTML]{ECF4FF}\textbf{0.052}} & \textbf{0.068} & \multicolumn{1}{c|}{\cellcolor[HTML]{ECF4FF}\textbf{97.83}} & \textbf{92.84} \\ \hline
\end{tabular}%
}
\vspace{-8pt}
\end{table}


\begin{figure}[!tb]
    \centering
	\includegraphics[width=3.4in]{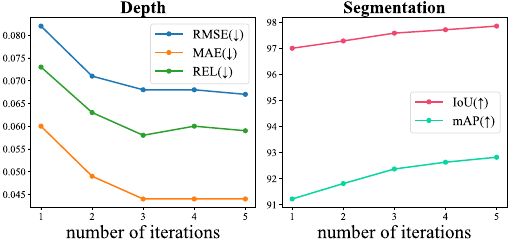}
    \caption{Ablation studies on the iterative strategy.}
    \label{fig_ablation}
    \vspace{-8pt}
\end{figure}

\subsection{Hybrid Loss Function}

Our proposed model is trained end-to-end using two loss functions for depth and segmentation.

{\bf Geometric Loss.} Following \cite{c8}, the depth estimation loss is formulated as:
\setlength{\arraycolsep}{0.0em}
\begin{eqnarray}
L_{geo} &{}={}& w_d {||D - D^*||}_2 + w_g {|| \nabla D - \nabla D^{*} ||_1} \nonumber \\
&& {+}\: w_n {||N_D - N_{D^*}||_1}
\end{eqnarray}
\setlength{\arraycolsep}{5pt}where the three terms represent the L2 loss between the predicted depth $D$ and the ground-truth depth $D^*$, and the L1 losses between the gradient and surface normal of $D$ and $D^*$, respectively. $w_d$, $w_g$ and $w_n$ are weights and are all set as 1 in practice.

{\bf Semantic Loss.} For semantic segmentation, the standard cross-entropy loss is used:
\begin{equation}
    L_{sem} = l_{ce} (S, S^*)
\end{equation}where $S$ and $S^*$ denote the predicted and ground-truth segmentation masks, respectively.

{\bf Overall}, the total loss function is formulated as:
\begin{equation}
    L = \alpha L_{geo} + \beta L_{sem}
\end{equation}
where $\alpha$ and $\beta$ are two hyper-parameters empirically set to 1 and 0.1 based on their relative magnitudes. The hybrid loss function is applied to multi-scale layers of the dual-branch decoder in each iteration.

\begin{figure*}[!tb]
    \centering
    \includegraphics[width=5in]{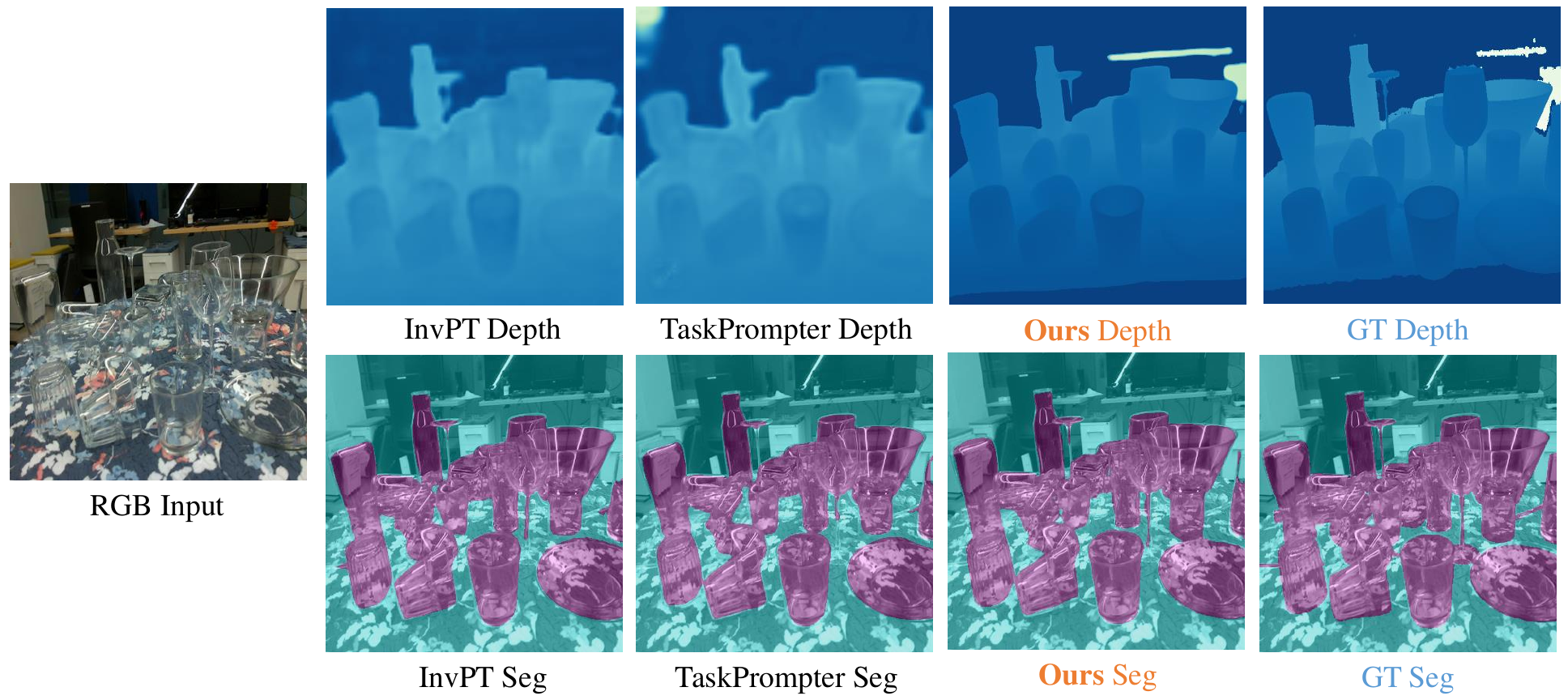}
    \caption{\textbf{Qualitative comparison on ClearPose dataset}, where Seg and GT stand for segmentation and ground truth respectively. Although this test scene is rather challenging, our method performs quite well in both depth and segmentation compared to SOTA multi-task methods.}
    \label{clearpose_fig}
\end{figure*}

\section{EXPERIMENTS}

\subsection{Experiment Setup}

{\bf Implementation Details.} Our model is implemented in PyTorch and trained on an RTX 4090 GPU with a batch size of 4 for 20 epochs. For all training, we use the Adam optimizer with a learning rate of 1e-5. The resolution of the input image is resized to $384 \times 384$, without using any image augmentation strategies such as random flipping or rotating.

{\bf Datasets.} To evaluate the effectiveness and robustness of our model, we conduct experiments on both synthetic dataset Syn-TODD \cite{c15} and real-world dataset ClearPose \cite{c41}. Syn-TODD is a photo-realistic dataset containing more than 113k image pairs with multi-task annotations, which is compatible with monocular, stereo, and multi-view methods. We follow the original paper \cite{c15} to prepare the dataset. ClearPose is a real-world dataset consisting of over 350k RGB-Depth frames. The dataset includes extreme scenarios such as heavy occlusions and non-planar configurations. We define two semantic categories, namely background and object, but note that it can be easily extended to other various classes. We follow the official setup to split the dataset into training and testing sets and the input depth map is not utilized.

{\bf Baselines.} Since our model is the first monocular multi-task framework for transparent objects, we compare it with two state-of-the-art stereo and multi-view methods elaborate for transparent objects, namely SimNet \cite{c14} and MVTrans \cite{c15}. The other two baselines are InvPT \cite{c42} and TaskPrompter \cite{c33}, which are designed for general multi-task dense predictions. SimNet takes stereo images as input, while MVTrans can be extended to 3 or 5 views. Both methods first construct a matching volume on the reference image through homography transformation, then perform multi-task predictions. InvPT and TaskPrompter are two state-of-the-art multi-task frameworks leveraging the interactions between different tasks with monocular images as input.

{\bf Evaluation Metrics.} For depth estimation, following \cite{c7}, root mean squared error (RMSE), absolute relative difference (REL), and mean absolute error (MAE)  are used as standard metrics. For semantic segmentation, we use intersection over union (IoU) and mean average precision (mAP) as \cite{c15} for fair comparison. IoU > 0.5 is used as the threshold in computing mAP.

\subsection{Comparison on Synthetic Dataset}

As shown in Table \ref{table_syntodd}, we conduct experiments against other baselines on the Syn-TODD dataset. The two monocular methods are reproduced using their default settings for fair comparison and the results of stereo and multi-view methods come from the original papers. Obviously, despite using only a single RGB image as input, our model significantly outperforms all other monocular and even multi-task baselines both in depth estimation and semantic segmentation, exceeding over 45\% on RMSE and REL compared to the second-best method. Besides, due to the lack of inter-task communication, other methods have unbalanced performance on different tasks. For example, monocular methods perform better on segmentation than depth, while multi-view methods do the opposite. In contrast, our method performs quite well on all tasks, which can be attributed to the cross-task fusion of the complementary information. 

Fig. \ref{syntodd_fig} shows a qualitative test example with transparent objects. As can be seen, since transparent objects usually inconsistently refract the background and lack texture in RGB images, the baseline methods tend to make unsatisfactory predictions, resulting in large areas of missing both in depth maps and segmentation masks. However, by iteratively exchanging useful semantic and geometric information, our method produces complete and sharp-edged results.

\subsection{Comparison on Real-world Dataset}

The quantitative results on the large-scale ClearPose dataset are shown in Table \ref{table_clearpose}. Since this dataset dose not support multi-view inputs, we only reproduce InvPT and TaskPrompter on ClearPose. Our approach consistently performs best on the real-world dataset, especially on the more challenging depth estimation task, affirming the robustness of our model.

In Fig. \ref{clearpose_fig}, visualizations of a complex test scene with heavy clutter and occlusion is provided to showcase the superiority of our method. Even humans find it difficult to accurately recognize and determine the geometric relations of each transparent object in this scene. In spite of achieving competitive results in 2D segmentation tasks, both InvPT and TaskPrompter lag far behind in depth estimation, producing blurry and noisy predictions. The results intuitively reveal that the multi-task baselines fail to utilize semantics to boost depth performance, and a single regression is not sufficient to obtain detailed results for transparent objects. However, our method overcomes these issues by gradually refining the multi-scale features, which allows the decoder to learn more details such as edges and surfaces.

\subsection{Ablation Studies}

{\bf Fusion Ablation.} In Table \ref{table_ablation}, we remove the semantic and geometric fusion module and retrain our model from scratch on Syn-TODD dataset, keeping other configurations the same. We also remove one of the depth and segmentation branches completely. The empirical results show that although predicting depth or segmentation alone is better than predicting them simultaneously, all three variants exhibit a significant drop in both depth and segmentation performance. The reason lies in that supervising the two branches independently without fusion would lead to conflict compared to single prediction, while the integration of our fusion module makes the gradient backpropagate between the two branches, which facilitates both tasks.

{\bf Iterative Strategy Ablation.} To investigate the utility of our iterative strategy, in Fig. \ref{fig_ablation}, we show the relationship between the number of iterations and the prediction performance of both tasks on Syn-TODD dataset. When the number of iterations is 1, the prediction process is equivalent to other models and the iterative strategy does not work. The supervision is gradually increased to ensure that the last iteration is fully supervised. It can be seen from the general trend of the curves that, as the number of iterations increases, the overall performance of both depth and segmentation improves accordingly, which demonstrates the effectiveness of our iteration strategy. The results reveal that only one regression produces suboptimal results, and by repeatedly updating the features, the model can be forced to gradually observe different details such as edges and surfaces, as humans do. Thus with an iterative strategy, our model can produce clearer dense predictions, which is especially useful for transparent objects. Considering the tradeoff between performance and memory footprint, we set the total number of iterations to 3.

\section{CONCLUSIONS}

In this work, we propose a monocular framework to jointly predict depth and segmentation for transparent objects. We present a semantic and geometric fusion module which can better leverage the complementary information of both tasks. An iterative strategy is also utilized to gradually refine the initial blurry results of untextured transparent objects. Experimental results demonstrate that our model outperforms state-of-the-art monocular and multi-view methods by a large margin, both on synthetic and real-world datasets.

\addtolength{\textheight}{-4cm}   




\section*{ACKNOWLEDGMENT}

This work is supported by the National Key Research and Development Program of China under Grant 2021ZD0114505, the Open Projects Program of State Key Laboratory of Multimodal Artificial Intelligence Systems under Grant No.MAIS2024112, and the Excellent Youth Program of State Key Laboratory of Multimodal Artificial Intelligence Systems.

\bibliographystyle{IEEEtran}
\bibliography{references}
\end{document}